\documentclass[preprint]{elsarticle}

\usepackage{amsmath, amssymb, amsthm}
\usepackage[utf8]{inputenc}
\usepackage[english]{babel}
\usepackage{enumitem}
\usepackage{graphicx}
\usepackage{hyperref}
\usepackage{booktabs}
\usepackage{mathrsfs}
\usepackage{algorithm}
\usepackage{algorithmic}

\newtheorem{remark}{Remark}[section]
\newtheorem{definition}[remark]{Definition}
\newtheorem{example}[remark]{Example}

\DeclareMathOperator{\POS}{\text{POS}}
\DeclareMathOperator{\att}{\mathcal{A}}
\DeclareMathOperator{\tnorm}{\mathcal{T}}

\DeclareMathOperator{\implicator}{\mathcal{I}}

\begin{document}

\begin{frontmatter}
    
\title{FRRI: a novel algorithm for fuzzy-rough rule induction}

\author[1]{Henri Bollaert}\corref{cor1}
\ead{henri.bollaert@ugent.be}
\author[1]{Marko Palangetić}
\author[1]{Chris Cornelis}
\author[2]{Salvatore Greco}
\author[3,4]{Roman Słowiński}
\cortext[cor1]{Corresponding author}

\affiliation[1]{
    organisation={Department of Applied Mathematics, Computer Science and Statistics, Ghent University},
    addressline = {Krijgslaan 281, S9},
    postcode={9000},
    city={Ghent},
    country={Belgium}
}
\affiliation[2]{
    organisation={Faculty of Economics, University of Catania},
    country={Italy}
}
\affiliation[3]{
    organisation={Institute of Computing Science, Poznań University of Technology},
    city={60-956 Poznań},
    country={Poland}
}
\affiliation[4]{
    organisation={Systems Research Institute, Polish Academy of Sciences},
    city={01-447 Warsaw},
    country={Poland}
}

\begin{abstract}
Interpretability is the next frontier in machine learning research. In the search for white box models --- as opposed to black box models, like random forests or neural networks --- rule induction algorithms are a logical and promising option, since the rules can easily be understood by humans. Fuzzy and rough set theory have been successfully applied to this archetype, almost always separately. As both approaches offer different ways to deal with imprecise and uncertain information, often with the use of an indiscernibility relation, it is natural to combine them. The QuickRules \cite{JensenCornelis2009} algorithm was a first attempt at using fuzzy rough set theory for rule induction. It is based on QuickReduct, a greedy algorithm for building decision superreducts. QuickRules already showed an improvement over other rule induction methods.
However, to evaluate the full potential of a
fuzzy rough rule induction algorithm, one needs to start from the foundations. Accordingly, the novel rule induction algorithm, Fuzzy Rough Rule Induction (FRRI), we introduce in this paper, uses an approach that has not yet been utilised in this setting. We provide background and explain the workings of our algorithm. Furthermore, we perform a computational experiment to evaluate the performance of our algorithm and compare it to other state-of-the-art rule induction approaches. We find that our algorithm is more accurate while creating small rulesets consisting of relatively short rules.
\end{abstract}

\begin{keyword}
    fuzzy rough set theory \sep rule induction \sep classification
\end{keyword}

\end{frontmatter}

\section{Introduction}\label{sec:intro}

Many fuzzy rule induction algorithms have been established, mostly for deriving a concise set of rules comprehensible by humans for tasks like classification and prediction. These include, for example, fuzzy association rule mining \cite{1593646, Dong2005}, first-order fuzzy rule generation \cite{DROBICS2003131, PradeRIchardSerr2003}, and linguistic semantics-preserving modeling \cite{MarnBlzquez2002FromAT, QIN2008435}.
However, the efficacy of most of the existing approaches to fuzzy rule induction is reduced as the data dimensionality increases. Some methods manage to avoid this, for example standard covering algorithms for rule induction (e.g., RIPPER \cite{COHEN1995115} and its fuzzy counterpart, FURIA \cite{FURIA}) that learn rules in an incremental way, with each rule in turn constructed by adding maximally informative features one by one.

QuickRules \cite{JensenCornelis2009} was developed to combat this problem by directly integrating the rule induction process in a greedy feature selection algorithm based on fuzzy rough set theory \cite{9908636, fuzzyroughsets, JensenShenbook2008}. Despite promising initial results reported in \cite{JensenCornelis2009}, the greedy nature of this algorithm does not allow it to outperform state-of-the-art algorithms w.r.t.\ balanced accuracy, ruleset size and rule length, as we will also demonstrate in the experimental section of this paper.

In this paper, we propose a novel rule induction algorithm called \emph{Fuzzy-Rough Rule Induction (FRRI)}. FRRI combines the best ingredients of fuzzy rule induction algorithms and rough rule induction algorithms to induce small sets of short rules that accurately summarize the training data and predict classes of new objects.

The remainder of this paper is structured as follows. 
Section \ref{sec:prelim} recalls the required theoretical background, including a short introduction on fuzzy set theory and rough set theory, and examines the existing work on fuzzy and rough rule induction. We give the explanation of the novel FRRI algorithm in Section \ref{sec:frri-expl}. Section \ref{sec:exp} describes the experiments that we set up to evaluate the performance of FRRI and to compare it to state-of-the-art fuzzy and rough set rule induction algorithms, in terms of prediction accuracy, number of generated rules and rule length. Finally, Section \ref{sec:conclusion} concludes the paper and outlines our future work.

\section{Preliminaries}\label{sec:prelim}

\subsection{Fuzzy sets}

In this paper, \(U\) will indicate a finite, non-empty set, called the \emph{universe}. A \emph{fuzzy set} \cite{fuzzysets} \(A\) in \(U\) is a function \(A \colon U \to [0,1]\). For \(u\) in \(U\), \(A(u)\) is called the membership degree of \(u\) in \(A\). The set of all fuzzy sets of $U$ is denoted as $\mathscr{F}(U)$. Let $A, B \in \mathscr{F}(U)$ be two fuzzy sets. $A$ is a fuzzy subset of $B$ if 
\[
(\forall u \in U)(A(u) \leq B(u))
\]
The union of a fuzzy set $A$ and a fuzzy set $B$ is the fuzzy set
\[
(\forall u \in U)((A \cup B)(u) = \max(A(u), B(u))
\]

In fuzzy logic and fuzzy set theory, we use \emph{triangular norms} and \emph{implicators} to generalize the classical connectives conjunction (\(\wedge\)) and implication (\(\rightarrow\)), respectively. A triangular norm or \emph{t-norm} is a function \(\tnorm \colon [0,1]^2 \to [0,1]\) which has \(1\) as neutral element, is commutative and associative and is increasing in both arguments. An \emph{implicator} is a function \(\implicator \colon [0,1]^2 \to [0,1]\) which is decreasing in the first argument and increasing in the second one. Additionally, the following boundary conditions should hold: 
\[\implicator(0,0) = \implicator(0,1) = \implicator(1,1) = 1 \text{ and } \implicator(1,0)=0.\]

A \emph{(binary) fuzzy relation} on a universe \(U\) is simply a fuzzy subset of \(U \times U\). A fuzzy relation \(R\) is \emph{reflexive} if \(R(u,u) = 1\) for all elements \(u \in U\). It is \emph{symmetric} if for every pair of elements \(u\), \(v\) in the universe, \(R(u, v) = R(v,u)\). For a given t-norm $\tnorm$, $R$ is \emph{$\tnorm$-transitive} if for every triplet of elements \(u, v, w\), $\tnorm(R(u, v), R(v, w)) \leq R(u, w)$. A reflexive and $\tnorm$-transitive relation is called a \emph{$\tnorm$-preorder} relation. A fuzzy \emph{tolerance} relation is a reflexive and symmetric fuzzy relation. Given an element \(u \in U\), its \emph{fuzzy tolerance class} w.r.t.\ to a tolerance relation \(R\) is the fuzzy set \(Ru (y) = R(u, v)\), for any \(v \in U\).

\subsection{Fuzzy rough sets}

In classical rough set theory \cite{roughsets}, we examine a universe divided into equivalence classes by an equivalence relation. We can then approximate a concept (i.e., a subset of \(U\)) using two crisp sets: the lower and the upper approximation. The \emph{lower approximation} contains all elements which are certainly part of the concept. It is defined as the union of the equivalence classes which are subsets of the concept. The \emph{upper approximation}, which contains all elements which might possibly be a part of the concept, is defined as the union of all equivalence classes which have a non-empty intersection with the concept in question. In fuzzy rough sets \cite{fuzzyroughsets}, we replace the crisp equivalence relation with a binary fuzzy relation \(R\) on \(U\) and, in turn, the upper and lower approximations also become fuzzy sets. Let \(A\) be a fuzzy set in \(U\). Furthermore, let \(\tnorm\) be a t-norm and \(\implicator\) be an implicator. We define the \emph{fuzzy rough set (FRS)} for the fuzzy set \(A\) w.r.t.\ the fuzzy relation \(R\) as the pair \((\underline{A}_R^{\implicator}, \overline{A}_R^{\tnorm})\), with, for \(u\) in \(U\):
\begin{itemize}
    \item The \emph{fuzzy lower approximation} \(\underline{A}_R^{\implicator}\) of $A$ is the fuzzy set
    \begin{equation}\label{eq:flowappr}        
        \underline{A}_R^{\implicator}(u) = \min_{v \in U} \implicator(R(u, v), A(v))
    \end{equation}
    \item The \emph{fuzzy upper approximation} \(\overline{A}_R^{\tnorm}\) is the fuzzy set
    \begin{equation}\label{eq:fuppappr}
        \overline{A}_R^{\tnorm}(u) = \max_{v \in U} \tnorm(R(u, v), A(v))
    \end{equation}
\end{itemize}

\subsection{Information and decision systems}\label{sec:infosys}

An \emph{information system} \cite{roughsets} is a pair \((U, \att)\), with \(U\)  a finite, non-empty \emph{universe} of objects, and \(\att\) a finite, non-empty set of \emph{attributes} describing these objects. Each attribute \(a \in \att\) has an associated domain of possible values \(V_a\) and an associated function \(a \colon U \to V_a\) which maps an object \(u \in U\) to its value \(a(u)\) for attribute \(a\). An attribute is \emph{numerical} if its value set is a closed real interval. 
In the remainder of this paper, we will only work in \emph{normalised} information systems with numerical condition attributes, where all attributes are scaled such that they have unit range, i.e., such that $V_a = [0,1]$ for all attributes $a \in \att$. This scaling step is important because the original range of a condition attribute often has no bearing on the true importance of the feature as a predictor for the class. To this aim, we will preprocess information systems by performing \emph{minimum-maximum normalisation}, i.e., we replace each original attribute function \(a \in \att\) with the following alternate attribute function $a'$:
\begin{align*}
	a' \colon &U \to [0, 1] \\
	&u \mapsto \frac{a(u) - \min \{a(v) \mid v \in U\}}{\max \{a(v) \mid v \in U\} - \min \{a(v) \mid v \in U\}}
\end{align*}
In Section \ref{sec:frri-expl}, we will sometimes look at novel objects that are not yet part of the information system under consideration. In such a case, it is possible that the value for a condition attribute $a$ of that object $v$ is outside of the range observed in the original information system. We solve this by setting $a'(v)$ to 1 if $a(v)$ is larger than $\max \{a(u) \mid u \in U\}$, and setting it to 0 if $a(v)$ is smaller than $\min \{a(u) \mid u \in U\}$.

In the following sections, we will assume each the condition attribute $a$ has been normalised, and stop writing $a'$.

If there is no inherent order on the values of an attribute, they can be compared by means of a binary fuzzy \emph{indiscernibility} relation \(R_i\) defined on \(U \times U\). In this paper, we will use the following relation
\begin{equation}\label{eq:indiscrel}
    R_i(u, v) = 1 - |a(u) - a(v)|
\end{equation}
However, imagine a case where an attribute represents length, which does have an inherent order. In such cases, we can use a fuzzy \emph{dominance} relation $R_d$, which is a fuzzy $\tnorm$-preorder relation for a given t-norm $\tnorm$ that encodes how much the first argument dominates the second. In this paper, we will use the following relation:
\begin{equation}
    R_d(u, v) = \min\left(1 - (a(v) - a(u)), 1\right)
\end{equation}
We can compare two objects in $U$ w.r.t.\ a subset \(B\) of \(\att\) using the fuzzy \(B\)-indiscernibility relation \(R_B\), defined as 
\begin{equation}
    R_B(u, v) = \tnorm(\underbrace{R_a(u, v)}_{a \in B})
\end{equation}
for each \(u\) and \(v\) in \(U\) and for a given t-norm \(\tnorm\), where $R_a$ is the relation used for comparing the values of attribute $a$. In general, \(R_B\) is a reflexive fuzzy relation.

A \emph{decision system} \((U, \att \cup \{d\})\) with a single \emph{decision attribute} \(d\) is an information system which makes a distinction between the \emph{condition} attributes \(\att\) and the decision attribute \(d\), which is not an element of \(\att\). In this paper, the decision attribute is assumed to be \emph{categorical}, which means that it has a finite, unordered value domain. We will call the values in \(V_d\), \emph{decision classes}. 
For a given object $u$ in $U$, we will identify its decision class $d(u)$ with the set of all objects that have the same value for the decision attribute.
Given a subset of attributes \(B \subseteq \att\), the fuzzy \(B\)-positive region is a fuzzy set in the universe \(U\) that contains each object \(u\) to the extent that all objects which are approximately similar to \(u\) w.r.t.\ the attributes in \(B\), have the same decision value \(d\) \cite{CORNELIS2010209}:
\[
\POS_B(u) = \left(\bigcup_{c \in V_d} \underline{c}_{R_B}^{\implicator} \right) (u)
\]
It is calculated as the membership degree of $u$ to the union of the fuzzy lower approximations of all decision classes of the decision system.

\subsection{Rule induction}\label{sec:RI-expl}

In the following sections, we will give a short description of some fuzzy and rough rule induction algorithms.
RIPPER \cite{COHEN1995115}, and its fuzzified version FURIA \cite{FURIA}, are state-of-the-art (fuzzy) rule induction algorithms that work in two phases. In the first phase, an initial ruleset is developed using a modified version of Incremental Reduced Error Pruning (IREP*) \cite{furnkranz1994incremental}. Afterwards, in the second phase, an optimisation is applied, which calculates possible alternatives for each rule, and selects the best one. To round out the optimisation step, IREP* is once again used to fill in any possible gaps left by selecting alternate rules, this optimisation step can be repeated multiple times.

Rough set theory is also widely applied to rule induction. An overview of applications of rough sets can be found here \cite{Chikalov2013}. Among rough set based rule induction algorithms, MODLEM \cite{Stefanowski1998OnRS} is generally considered to be the standard option due to its excellent performance. This algorithm was inspired by LEM2 \cite{threediscretmodlem}. By contrast to the former, it has no need for a discretisation step, selecting the best cut-off values for numerical attributes during the rule generation. MODLEM follows a heuristic strategy for creating an initial rule by choosing sequentially the ``best'' elementary conditions according to some heuristic criteria.
Learning examples that match this rule are removed from consideration. The process is repeated iteratively as long as some learning examples remain uncovered. The resulting set of rules covers all learning examples.

There have been very few attempts at developing rule induction algorithms based on fuzzy rough set theory. A prominent example is QuickRules \cite{JensenCornelis2009}, a greedy fuzzy rough algorithm based on the feature selection algorithm QuickReduct \cite{fuzzyroughsets, JensenShenbook2008}. It greedily selects the most discerning attribute according to its contribution to the positive region. During these calculations, candidate rules are checked and possibly added to a growing ruleset.
Recently, the use of the Hamacher t-norm in the context of QuickRules has been explored in \cite{hadamacher}.

Other work has largely focused on using classical rough set theory to generate fuzzy rulesets \cite{Hsiesh2008, ShenChouchoulas02}, but mainly ignores the direct use of fuzzy-rough concepts.
The induction of gradual decision rules, based on fuzzy-rough hybridisation, has been presented in \cite{GRECO2006179}.
For this approach, new definitions of fuzzy lower and upper approximations were constructed that avoid the use of fuzzy logical connectives altogether. In this approach, decision rules are induced from lower and upper approximations defined for positive and negative relationships between credibility of premises and conclusions. Only the ordinal properties of fuzzy membership degrees are used.
Another fuzzy-rough approach to fuzzy rule induction was presented in \cite{WangTsangZhaoChen2007},
where fuzzy reducts are employed to generate rules from data. This method also employs a fuzzy-rough feature selection preprocessing step.
In \cite{JensenShenbook2008}, a fuzzy decision tree algorithm was proposed, based on fuzzy ID3, that incorporates the fuzzy-rough dependency function as a splitting criterion.
A fuzzy-rough rule induction method was also proposed in \cite{Hong2006} for generating certain and possible rulesets from hierarchical data. This differs from our approach, in which we will only generate a single ruleset, as we will explain in the next section.

\section{FRRI: fuzzy-rough rule induction}\label{sec:frri-expl}

Let \((U, \att \cup \{d\})\) be a decision system, where the universe \(U\) is a finite set of objects, \(\att = \{a_1, a_2, \dots, a_m\}\), for a natural number $m$, is the set of (numerical) condition attributes,
and \(d\) is the (categorical) decision attribute. We will also refer to this decision system as the training set. The condition attributes are normalised (on the training set) to have a range of \([0,1]\), as explained in Section \ref{sec:infosys}. The ordering defined on the set of attributes is fixed but arbitrary, as it is often given with real world data. The FRRI algorithm constructs fuzzy decision rules from this decision system, which can be used to summarize the knowledge contained in the data, and to make predictions on the decision attribute of new, unseen objects that are not in the training set. We will define the format of these rules in Section \ref{sec:ruleformat}.

In FRRI, each rule in the final ruleset corresponds to an original object in $U$. Our algorithm consists of two basic steps:
\begin{enumerate}
    \item Rule shortening: in this step, we look at the initial rules derived from each object of the training set. For each of those rules, we discard conditions, as long as they do not decrease the discerning power of that object. This attribute reduction is performed by constructing a new comparison relation. The shortened rules form an initial ruleset.
    \item Rule selection: in this step, we select a minimal number of (shortened) rules from the initial ruleset which still cover the entire training set. This is done by solving an integer programming problem. An optimal solution of this problem defines the final ruleset.
\end{enumerate}
We will now explain each of these steps in detail, but first we will define the rule format used in FRRI.

\subsection{Rule format}\label{sec:ruleformat}
A rule $r$ in FRRI has the general form 
    \begin{quote}
        \textsc{if} antecedent \textsc{then} consequent
    \end{quote}
with the antecedent and the consequent expressed in terms of fuzzy membership. Each rule is originally associated with an object $u_r$ from the training set.
We will now take a closer look into the exact format of the rule antecedent and consequent.
\begin{definition}
    The \emph{antecedent} of a rule $r$ is a conjunction of conditions, each of them associated with attribute \(a\) in \(\att\). Consider now such attribute $a$ and object $u_r$ from $U$. We consider three types of conditions:
    \begin{itemize}
        \item \textsc{similar} condition: \(a\) is similar to the value \(a(u_r)\)
        \item \textsc{dominant} condition: \(a\) is smaller than or similar to the value \(a(u_r)\)
        \item \textsc{dominated} condition: \(a\) is greater than or similar to the value \(a(u_r)\)
    \end{itemize}
\end{definition}
The antecedent of a rule is then written as a set of (unordered) conditions, linked with \textsc{and}. Note that not all attributes need to be present in a given rule. For the rule $r$, we encode the type of condition associated with each attribute in a type mapping:
\begin{equation*}\label{eq:type-mapping}
\textsc{type}_r: \att \to \{\textsc{similar, dominant, dominated, unused}\}
\end{equation*}
where \textsc{unused} designates that the attribute is not used in that rule.

We look for minimal rules, that is, rules from which no condition attribute can be removed. Moreover, we prefer \textsc{dominant} or \textsc{dominated} conditions over \textsc{similar}, since the former are more general than the latter. The rule shortening step, explained in the next section, is designed to incorporate this preference.
The antecedent of a rule induces a fuzzy set, which we can use to calculate the matching degree of an object from the universe to that rule.
\begin{definition}
    Let $r$ be a rule derived from object $u_r$. Let $v$ be an object in $U$, and let $\tnorm$ be a t-norm. The \emph{matching degree} of $v$ to $r$ is defined as
    \begin{equation*}
            M_r(v) = \min\left\langle R_{\textsc{type}_r(a)}(a(u_r), a(v)) \mid a \in \att \right\rangle
    \end{equation*}
    We also denote the fuzzy set $M_r$ as the \emph{matching set} of $r$.
    The relations corresponding to each condition type are, for $u, v \in [0,1]$:
    \begin{align*}
        R_\textsc{similar} (u, v) &= R_i(u, v)\\
        R_\textsc{dominant} (u, v) &= R_d(u, v) \\
        R_\textsc{dominated} (u, v) &= R_d(v, u) \\
        R_\textsc{unused} (u, v) &= 1
    \end{align*}
\end{definition}
We can see that \textsc{dominant} and \textsc{dominated} conditions are indeed more general than \textsc{similar} conditions, since, for the same $u \in [0,1]$, it holds that, for any $v \in [0,1]$,
\begin{equation*}
    R_\textsc{similar} (u, v) \leq R_\textsc{dominant}(u, v) \text{ and } R_\textsc{similar} (u, v) \leq R_\textsc{dominated}(u, v)
\end{equation*}
 
\begin{definition}
    Let $\implicator$ be an implicator. The \emph{consequent} of a rule $r$ derived from object $u_r$, is written as: ``$d$ is $d(u_r)$''. It is encoded as a fuzzy singleton in $V_d$:
    \begin{equation*}
        \{ (d(u_r), \underline{d(u_r)}^{\implicator}_{R_{\att}}(u_r) ) \}
    \end{equation*}
\end{definition}
This membership degree to the lower approximation of its decision class (Equation \eqref{eq:flowappr}) represents the degree to which object $u_r$ is consistent with its decision class, and can be seen as the confidence score of the rule.

We now have all the information required to define the degree to which a rule covers an object.

\begin{definition}\label{def:covering}
    Consider a rule $r$ derived from object $u_r$ and another object $v$. The rule \emph{$r$ covers $v$ to the degree}
    \begin{equation*}
        S_r(v) = \min\left(M_r(v), \underline{d(u_r)}^{\implicator}_{R_{\att}}(u_r)\right)
    \end{equation*}
    Moreover, we say that the rule $r$ \emph{covers} object $v$ if $S_r(v) > 0$, and we call the fuzzy set $S_r$ the \emph{covering set} of $r$.
\end{definition}

\begin{table}[ht]
    \centering
    \begin{tabular}{llllllllll}
    \toprule
    & $a_1$ & $a_2$ & $a_3$ & $a_4$ & $a_5$ & $a_6$ & $a_7$  & $a_8$ & $d$ \\ 
     \midrule
    $u_1$ & 0.12 & 0.07 & 0.00 & 0.04 & 0.00 & 0.17 & 0.45 & 0.11 & 0 \\
    $u_2$ & 1.00 & 0.88 & 1.00 & 0.45 & 0.86 & 0.52 & 0.37 & 1.00 & 1 \\
    $u_3$ & 0.88 & 0.60 & 0.40 & 0.62 & 1.00 & 0.56 & 0.69 & 0.56 & 0 \\
    $u_4$ & 0.88 & 1.00 & 0.45 & 0.55 & 0.88 & 0.67 & 0.10 & 0.53 & 1 \\
    $u_5$ & 0.00 & 0.05 & 0.95 & 1.00 & 0.24 & 1.00 & 1.00 & 0.25 & 0 \\
    $u_6$ & 0.00 & 0.11 & 0.35 & 0.60 & 0.35 & 0.81 & 0.00 & 0.00 & 0 \\
    $u_7$ & 0.12 & 0.00 & 0.40 & 0.00 & 0.01 & 0.00 & 0.20 & 0.08 & 0 \\
    \bottomrule
    \end{tabular}
    \caption{Didactic example used throughout the explanation of this algorithm.}
    \label{tab:og-ex}
\end{table}

\begin{example}
    Throughout this section, we will use the decision system presented in Table \ref{tab:og-ex} to illustrate the operation of our algorithm. It is a sample of the Pima Indians Diabetes dataset taken from the KEEL dataset repository \cite{keeldatarepo}, which we have normalised using min-max normalisation, such that all attributes have unit range. We consider a universe with seven objects, eight numerical condition attributes and one binary decision attribute.
    We consider an example rule $r$ derived from object $u_1$ from our example decision system:
    \begin{quote}
        \textsc{if} $a_1$ is similar to $0.12$ \textsc{and} $a_2$ is greater than or similar to $0.07$ \textsc{and} $a_7$ is smaller than or similar to $0.45$ \textsc{then} $d$ is $0$.
    \end{quote}
    The matching degree of $u_2$ to this rule is:
    \begin{align*}
        M_r(u_2) 
        &= \min(R_\textsc{similar}(0.12, 1.00), R_\textsc{dominated}(0.07, 0.88), R_\textsc{dominant}(0.45, 0.37)) \\
        &= \min( \begin{aligned}[t]
            &1 - |0.12-1|, \\
            &\min\left(1 - (0.07 - 0.88), 1\right), \\
            &\min\left(1 - (0.37 - 0.45), 1\right))
        \end{aligned} \\
        &= \min(0.12, 1, 1)\\
        &= 0.12
    \end{align*}
    The degree to which this rule covers $u_2$ is
    \begin{equation*}
        S_r(u_2) = \min(M_r(u_2), \underline{d(u_r)}^{\implicator}_{R_{\att}}(u_r)) = \min(0.12, 0.934783) = 0.12
    \end{equation*}
\end{example}

\subsection{Rule shortening}

As mentioned before, in FRRI, each object of the training set is the starting point of a rule.

\begin{definition}
    Consider an object $u$ in the training set. The \emph{total rule} $r(u)$ corresponding to $u$ is the rule
    \begin{align*}
        \textsc{if} \quad &a_1 \text{ is similar to } a_1(u) \; \textsc{and}\\
        &a_2 \text{ is similar to } a_2(u) \; \textsc{and}\\
        &\dots \\
        &a_m \text{ is similar to } a_m(u)\\
        \textsc{then} \quad &d \text{ is } d(u)
    \end{align*}
\end{definition}

The rule shortening step prunes this total rule of each object to a final shortened, more general rule $\hat{r}(u)$ in the procedure \textsc{rule\_prune}. The pseudo-code for this procedure can be found in Algorithm \ref{alg:rule_prune}. It is repeated independently for each object in the training set.
\textsc{Rule\_prune} iteratively tries to generalize each condition of $r(u)$ without making the new rule cover objects outside of the class indicated in its consequent. More formally, consider an attribute $a_i$ for an object $u$ (line $5$ of Algorithm \ref{alg:rule_prune}). We consecutively examine the possibility of setting the type of the condition corresponding to $a_i$ for the rule corresponding to $u$ as \textsc{unused, dominant, dominated} or \textsc{similar}. If for a type $t$, $S_{r^*(u)}$ is a subset of $d(u)$, then we set the type of $a_i$ to that type $t$ for this object. If not, we continue to the next type in the list.
We repeat this process for each attribute in the order $a_1, a_2, \dots, a_n$, and we repeat this \textsc{rule\_prune} procedure for each object in the training set. The result is a shortened, initial ruleset
\[\widehat{\mathscr{R}} = \{\hat{r}(u) \mid u \in U\}\]

\begin{algorithm}[htp]
\caption{The \textsc{rule\_prune} procedure}
\label{alg:rule_prune}
\begin{algorithmic}[1]
\STATE \textbf{Input:} object $u \in U$
\STATE \textbf{Output:} shortened rule $\hat{r}(u)$
\STATE $t^* = (t^*_1, t^*_2, \dots, t^*_m) \leftarrow (\textsc{similar}, \textsc{similar}, \dots, \textsc{similar})$
\FOR{$i=1$ \TO $m$}
    \FORALL{$t \in \{\textsc{unused, dominant, dominated, similar}\}$}
    \STATE $t^*_i \leftarrow t$
    \STATE $r^*(u) \leftarrow r(u)$ with $t^*$ as type vector
    \IF{$S_{r^*(u)} \subseteq d(u)$}
    \STATE break for-loop
    \ENDIF
    \ENDFOR
\ENDFOR
\RETURN $\hat{r}(u) \leftarrow r^*(u)$
\end{algorithmic}
\end{algorithm}

\begin{example}
    Let us apply this rule shortening step to object $u_1$ of our example dataset.
    Consider the first attribute $a_1$. We calculate $S_{r^*(u_1)}$ where $r^*(u_1)$ uses the following type mapping:
    \[
    (\textsc{unused, similar, similar, similar, similar, similar, similar})
    \]
    We should do this only for those objects from a different class, since for objects $v$ in $d(u_1)$ it is obvious that 
    \[
    S_{r^*(u_1)}(v) \leq (d(u_1))(v) = 1
    \]
    We find:
    \begin{align*}
        S_{r^*(u_1)}(u_2) &= \min\left(\min_{2 \leq i \leq 8}(R_{\textsc{similar}}(a_i(u_1), a_i(u_2)), \underline{d(u_1)}^{\implicator}_{R_{\att}}(u_1)\right) \\
        &= 0 \\
        S_{r^*(u_1)}(u_4) &= \min\left(\min_{2 \leq i \leq 8}(R_{\textsc{similar}}(a_i(u_1), a_i(u_4)), \underline{d(u_1)}^{\implicator}_{R_{\att}}(u_1)\right) \\
        &= 0
    \end{align*}
    As such, we can see that $S_{r^*(u_1)} \subseteq d(u_1)$, and we fix the type of $a_1$ for $u_1$ as \textsc{unused}.
    Now we move on to attribute $a_2$, and we try the following type mapping:
    \[
    (\textsc{unused, unused, similar, similar, similar, similar, similar})
    \]
    Again we calculate $S_{r^*(u_1)}$ with this type mapping, and we find:
    \begin{align*}
        S_{r^*(u_1)}(u_2) &= \min\left(\min_{3 \leq i \leq 8}(R_{\textsc{similar}}(a_i(u_1), a_i(u_2)), \underline{d(u_1)}^{\implicator}_{R_{\att}}(u_1)\right) \\
        &= 0 \\
        S_{r^*(u_1)}(u_4) &= \min\left(\min_{3 \leq i \leq 8}(R_{\textsc{similar}}(a_i(u_1), a_i(u_4)), \underline{d(u_1)}^{\implicator}_{R_{\att}}(u_1)\right) \\
        &= 0.058966
    \end{align*}
    This means we cannot set the type of $a_2$ to \textsc{unused} for $u_1$. When we calculate $S_{r^*(u_1)}$ with the mapping 
    \[
    (\textsc{unused, dominant, similar, similar, similar, similar, similar}),
    \]
    we do find that it is a subset of $d(u_1)$, and we fix this type as well.
    We continue this for the remainder of the attributes, until we arrive at the final type mapping of $u_1$:
    \[
    \textsc{type}_{x_1}(a) = \begin{cases}
        \textsc{dominant} & \text{if } a \in \{a_2, a_3\}, \\
        \textsc{unused} & \text{otherwise.}
    \end{cases}
    \]
    This results in the following shortened rule $\hat{r}(u_1)$:
    \begin{quote}
        IF $a_2$ is smaller than or similar to 0.07 AND $a_3$ is smaller than or similar to $0$ THEN $d$ is $0$.
    \end{quote}
\end{example}

\subsection{Rule selection}\label{sec:rule_selec}

After we have finished the rule shortening step, we will select a subset of rules from $\widehat{\mathscr{R}}$ to obtain the final ruleset $\mathscr{R}$. Using Definition \ref{def:covering}, we introduce the notation $z_{u,v}$ and the binary variable $r_u$ for all objects $u, v \in U$:
\begin{equation*}
    z_{u, v} = \begin{cases}
        1 & \text{if } \hat{r}(u) \text{ covers } v \\
        0 & \text{otherwise}
    \end{cases} 
    \; \text{  and  } \;
    r_u = \begin{cases}
        1 & \text{if } \hat{r}(u) \in  \mathscr{R}\\
        0 & \text{otherwise}
    \end{cases}
\end{equation*}
Finally, we can formulate the optimisation problem as follows: 
\[\text{Minimize: }\sum_{u \in U} r_u\]
subject to the conditions
\[(\forall v \in U)\left(\sum_{u \in U}r_u \times z_{u, v} \geq 1 \right),\]
or in other words: minimize the number of selected rules, while still assuring that every object in the universe is covered by at least one of the selected rules. This integer programming problem can be solved with classical techniques, such as branch-and-bound algorithms.

\begin{algorithm}[hbt]
    \caption{The \textsc{rule\_selection} procedure}
    \label{alg:rule-selection}
    \begin{algorithmic}[1]
    
        \STATE \textbf{Input:} universe $U$, initial ruleset $\widehat{\mathscr{R}}$
        \STATE \textbf{Output:} final ruleset $\mathscr{R}$
        
        \FORALL{$u$ in $U$}
            \FORALL{$v$ in $U$}
                \IF{$S_{\hat{r}(u)}(v) > 0$}
                    \STATE $z_{u, v} \leftarrow 1$
                \ELSE
                    \STATE $z_{u, v} \leftarrow 0$
                \ENDIF
            \ENDFOR
            \STATE \textbf{define variable} $r_u$ is 1 iff $\hat{r}(u) \in \mathscr{R}$ 
        \ENDFOR
        \STATE \textbf{Minimize} $\sum\limits_{u\in U} r_u$ such that $(\forall v \in U)\left(\sum\limits_{u \in U}r_u \times z_{u, v} \geq 1 \right)$
        \STATE $\{r_u^* \mid u \in U\} \leftarrow$ solution of above problem
        \RETURN $\{\hat{r}(u) \mid u \in U \text{ and } r_u^* = 1\}$
    \end{algorithmic}
\end{algorithm}

\begin{table}[ht]
    \centering
    \begin{tabular}{c|ccccccc}
         & $u_1$ & $u_2$ & $u_3$ & $u_4$ & $u_5$ & $u_6$ & $u_7$ \\
         \hline
        $u_1$ & 1 & 0 & 1 & 0 & 0 & 1 & 1 \\
        $u_2$ & 0 & 1 & 0 & 1 & 0 & 0 & 0 \\
        $u_3$ & 1 & 0 & 1 & 0 & 1 & 0 & 1 \\
        $u_4$ & 0 & 1 & 0 & 1 & 0 & 0 & 0 \\
        $u_5$ & 1 & 0 & 1 & 0 & 1 & 1 & 1 \\
        $u_6$ & 1 & 0 & 1 & 0 & 1 & 1 & 1 \\
        $u_7$ & 1 & 0 & 1 & 0 & 1 & 1 & 1
    \end{tabular}
    \caption{The variables $z_{u,v}$ for our example dataset.}
    \label{tab:coveringvarexample}
\end{table}

\begin{example}\label{ex:rule-selection}
    First, we calculate the values of $z_{u, v}$ for our example dataset in Table \ref{tab:coveringvarexample}.
    We can see that no single rule covers the entire training set. However, the following ruleset does:
    \[
    \mathscr{R} = \{\hat{r}(u_2), \hat{r}(u_5)\}
    \]
    $\mathscr{R}$ results in a minimal ruleset that still covers the entire training set.
    The corresponding rules are:
    \begin{quote}
        \textsc{if} $a_3$ is greater than or similar to 1 \textsc{and} $a_7$ is greater than or similar to 0.37 \textsc{then} $d$ is 1 \\
        \textsc{if} $a_1$ is smaller than or similar to 0 \textsc{and} $a_2$ is smaller than or similar to 0.05 \textsc{then} $d$ is 0
    \end{quote}
\end{example}
In general, the optimal solution is not univocal.

\subsection{Inference step}

Now, consider a new object \(v\) for which we want to predict the decision attribute given its condition attributes and a ruleset \(\mathscr{R}\) created by algorithm \ref{alg:rule-selection}. 
We derive the inference result as follows.
For each possible value \(c\) of the decision attribute (i.e., for each class), we calculate the confidence score as the maximal covering degree of $z$ to a rule whose consequent matches class $c$:
\[s_c = \max \{S_{r}(z) \mid r \in \mathscr{R} \text{ and } d(u_r) = c\}\]
and we assign to \(z\) the class with the highest confidence score:
\[\arg \max_{c \in V_d} s_c\]

\begin{example}
    Consider a test sample 
    \[v = (0.12, 0.40, 0.45, 0.48, 0.00, 0.48, 0.30, 0.50)\]
    We use the ruleset we found in Example \ref{ex:rule-selection}. The confidence score of class 0 is
    \begin{align*}
        s_0 &= S_{r(u_5)}(v) \\
        &= \min(R_{\textsc{dominant}}(0, 0.12), R_{\textsc{dominant}}(0.05, 0.40), 0.945652) \\
        &= \min(0.78, 0.35, 0.945652) = 0.35
    \end{align*}
    and the confidence score of class 1 is
    \begin{align*}
        s_1 &= S_{r(u_5)}(v) \\
        &= \min(R_{\textsc{dominated}}(1, 0.45), R_{\textsc{dominated}}(0.37, 0.30), 0.600000) \\
        &= \min(0.55, 0.93, 0.600000) = 0.55
    \end{align*}
    Therefore, the predicted class of $v$ would be $1$.
\end{example}

\section{Experimental evaluation}\label{sec:exp}

\subsection{Methods and evaluation measures}

\begin{table}[ht]
\centering
\begin{tabular}{llrrrr}
\toprule
dataset         & abbr.        & \(|\att|\) & \(|U|\) & \(|V_d|\) & IR    \\ \midrule
australian      & aus          & 14      & 690     & 2         & 1.25  \\
bands           & bands        & 19      & 365     & 2         & 1.70  \\
bupa            & bupa         & 6       & 345     & 2         & 1.38  \\
cleveland       & cleve        & 13      & 297     & 5         & 12.62 \\
dermatology     & derma        & 34      & 358     & 6         & 5.55  \\
ecoli           & ecoli        & 7       & 336     & 8         & 28.60 \\
glass           & glass        & 9       & 214     & 6         & 8.44  \\
heart           & heart        & 13      & 270     & 2         & 1.25  \\
ionosphere      & iono         & 33      & 351     & 2         & 1.79  \\
pima            & pima         & 8       & 768     & 2         & 1.87  \\
sonar           & sonar        & 60      & 208     & 2         & 1.14  \\
spectfheart     & spect        & 44      & 267     & 2         & 38.56 \\
vehicle         & vehi         & 18      & 846     & 4         & 1.10  \\
vowel           & vowel        & 13      & 990     & 11        & 1.00  \\
wine            & wine         & 13      & 178     & 3         & 1.48  \\
winequality-red & red          & 11      & 1599    & 6         & 68.10 \\
wisconsin       & wisc         & 9       & 683     & 2         & 1.86  \\
yeast           & yeast        & 8       & 1484    & 10        & 92.60 \\ \bottomrule
\end{tabular}
\caption{Details of the benchmark datasets used for these experiments: abbreviated name, the amount of features of each dataset ($|\att|$), the number of instances ($|U|$) and classes ($|V_d|$), and the imbalance ratio IR}
\label{tab:datasets}
\end{table}

We perform a computational experiment to evaluate the performance of our novel FRRI algorithm on 18 benchmark numerical datasets from the KEEL dataset repository\footnote{Some of these datasets usually have categorical condition features. We opted to remove these features from the dataset before using them in our experiments.} \cite{keeldatarepo}. The details of these datasets can be found in Table \ref{tab:datasets}, where we list the amount of features of each dataset ($|A|$), as well as the number of instances ($|U|$) and classes ($|V_d|$). Finally, the table also contains the imbalance ratio (IR) of each dataset, which is calculated as the size of the largest class divided by the size of the smallest class. We also list an abbreviated name for each dataset.

We implement our algorithm in Python, using the \textsc{gurobipy} package for Python \cite{gurobi} to solve the optimisation problem in the rule selection step.
We chose Gurobi because it is a mature, well tested product with good support, free access for academic use, with a strong community and an excellent Python implementation. However, any other exact optimizer (including open-source alternatives) could be used
without fundamentally changing the FRRI algorithm and the obtained experimental results, which were obtained after repeated runs.

We compare FRRI with QuickRules \cite{JensenCornelis2009}, MODLEM \cite{Stefanowski1998OnRS}, FURIA \cite{FURIA} and RIPPER \cite{COHEN1995115}. We use the implementation found in WEKA \cite{WEKA} for the latter algorithms and we use our own Python implementation of QuickRules, which was verified against the one in WEKA. We also use the default parameters for all algorithms.

For each algorithm, we perform ten-fold cross-validation on each dataset, using the same folds for each algorithm, after which we calculate the average balanced accuracy of the algorithm on the ten training folds. The balanced accuracy is the arithmetic mean of the recall on each class. Moreover, we also calculate the average number of rules and the average rule length over each ruleset and over each of the ten folds of each dataset. To determine the statistical significance of our results, 
we apply a two-stage procedure, which starts with a Friedman test \cite{multiple-comp-friedman}. If the result of that test is significant, we continue with a Conover post-hoc pairwise comparison procedure \cite{multiple-comp-conover, conover_test_big_citations} to determine the location of the significant differences. Moreover, we also use the Wilcoxon test \cite{pairwise-comp-wilcoxon} to detect differences between two methods.

\subsection{Results}

\begin{table}[bhpt]
    \centering
    \begin{tabular}{lrrrrr}
    \toprule
    dataset         &            FRRI &          MODLEM &      QuickRules &           FURIA &          RIPPER \\
    \midrule
    aus             &           0.788 &           0.861 &  \textbf{0.865} &  \textbf{0.865} &           0.852 \\
    bands           &  \textbf{0.675} &           0.604 &           0.533 &           0.629 &           0.582 \\
    bupa            &           0.601 &           0.652 &           0.500 &  \textbf{0.657} &           0.653 \\
    cleve           &  \textbf{0.290} &           0.276 &           0.208 &           0.245 &           0.238 \\
    derma           &           0.895 &           0.941 &           0.522 &  \textbf{0.950} &           0.942 \\
    ecoli           &  \textbf{0.721} &           0.512 &           0.176 &           0.530 &           0.545 \\
    glass           &  \textbf{0.679} &           0.518 &           0.314 &           0.522 &           0.495 \\
    heart           &           0.767 &           0.760 &           0.786 &  \textbf{0.818} &           0.762 \\
    iono            &  \textbf{0.912} &           0.890 &           0.659 &           0.887 &           0.892 \\
    pima            &           0.688 &           0.699 &           0.503 &  \textbf{0.703} &           0.699 \\
    sonar           &           0.786 &           0.705 &           0.523 &  \textbf{0.800} &           0.705 \\
    spect           &  \textbf{0.629} &           0.602 &           0.559 &           0.557 &           0.578 \\
    vehicle         &           0.629 &  \textbf{0.740} &           0.385 &           0.718 &           0.688 \\
    vowel           &  \textbf{0.912} &           0.684 &           0.163 &           0.833 &           0.705 \\
    wine            &           0.921 &  \textbf{0.961} &           0.916 &           0.950 &           0.935 \\
    red             &           0.333 &  \textbf{0.339} &           0.182 &           0.155 &           0.150 \\
    wisc            &           0.939 &  \textbf{0.955} &           0.933 &           0.952 &           0.953 \\
    yeast           &           0.484 &           0.454 &           0.125 &           0.544 &  \textbf{0.550} \\
    \midrule
    mean            &  \textbf{0.703} &           0.675 &           0.492 &           0.684 &           0.662 \\
    average rank    &            2.56 &            2.72 &            4.44 &   \textbf{2.28} &            3.00 \\
    \bottomrule
    \end{tabular}
    \caption{Balanced accuracy of FRRI, MODLEM, QuickRules, FURIA and \mbox{RIPPER} on the benchmark datasets.}
    \label{tab:first-acc}
\end{table}

Table \ref{tab:first-acc} contains the average balanced accuracy of these five algorithms on the benchmark set of datasets, Table \ref{tab:first-rules} contains the average number of rules, and Table \ref{tab:first-atts} the average rule length. In each table, the best result for each dataset is highlighted in bold.

Starting with the balanced accuracy, we can see that FRRI is on average the best performing model, beating the state-of-the-art algorithms on average by a comfortable margin. The Friedman test detected a significant difference ($p < 0.001$) between the models. The post-hoc procedure identified that QuickRules performed significantly worse than all other algorithms. We can see that FRRI's accuracy is never more than $0.11$ lower than the best accuracy on that dataset, while this difference is up to around $0.20$ for MODLEM, FURIA and RIPPER and over $0.74$ for QuickRules. Moreover, of all the algorithms, FRRI is most often the best performer. Finally, when we look at the average rank of each algorithm, we see that FURIA has the lowest average rank, closely followed by FRRI and MODLEM.

Interestingly, when we look at highly imbalanced datasets (IR $> 6$), we find that FRRI consistently and significantly ($p < 0.05$) outperforms the other algorithms, indicating that our algorithm is able to handle such datasets well.

\begin{table}[hpt]
    \centering
    \begin{tabular}{lrrrrr}
    \toprule
    dataset         &            FRRI &          MODLEM &      QuickRules &           FURIA &          RIPPER \\
    \midrule
    aus             &              92 &    121 &   732 &   \textbf{6} &   \textbf{6} \\
    bands           &              59 &    113 &   223 &           19 &   \textbf{4} \\
    bupa            &              77 &    103 &   332 &           12 &   \textbf{5} \\
    cleve           &             102 &     95 &   331 &            7 &   \textbf{4} \\
    derma           &              20 &     27 &    90 &           11 &   \textbf{8} \\
    ecoli           &              57 &     56 &   315 &           15 &   \textbf{9} \\
    glass           &              51 &     50 &   226 &           13 &   \textbf{8} \\
    heart           &              45 &     62 &   304 &           15 &   \textbf{4} \\
    iono            &              19 &     30 &   460 &           14 &   \textbf{7} \\
    pima            &             125 &    191 &   825 &           13 &   \textbf{4} \\
    sonar           &              18 &     63 &   276 &           10 &   \textbf{5} \\
    spect           &              24 &     55 &   126 &           17 &   \textbf{3} \\
    vehicle         &             162 &    177 &   603 &           31 &  \textbf{17} \\
    vowel           &             123 &    200 &  1062 &           69 &  \textbf{42} \\
    wine            &               8 &     13 &   211 &            7 &   \textbf{5} \\
    red             &             399 &    376 &  1541 &           28 &  \textbf{14} \\
    wisc            &              23 &     31 &   280 &           15 &   \textbf{5} \\
    yeast           &             485 &    337 &  1408 &  \textbf{19} &           21 \\
    \midrule
    mean            &             105 &    117 &   519 &           18 &  \textbf{10} \\
    \bottomrule
    \end{tabular}
    \caption{The average number of rules of FRRI, MODLEM, QuickRules, FURIA and RIPPER on the benchmark datasets.}
    \label{tab:first-rules}
\end{table}

\begin{table}[hpt]
\centering
\begin{tabular}{@{}lrrrr@{}}
\toprule
           & FRRI    & MODLEM  & QuickRules & FURIA     \\ \midrule
MODLEM     & $0.407$ &  &     &   \\
QuickRules & $0.009$ & $0.057$ & &   \\
FURIA      & $0.051$ & $0.007$ & $<0.001$ &   \\
RIPPER     & $<0.001$ &$ <0.001$ & $<0.001$ & $0.244$ \\ \bottomrule
\end{tabular}
\caption{The $p$-values obtained by performing the Conover post-hoc pairwise comparison procedure after a significant Friedman test on the ruleset sizes of FRRI, MODLEM, QuickRules, FURIA and RIPPER on the benchmark datasets.}
\label{tab:rules-friedman}
\end{table}

Let us now look at the  average size of the rulesets generated by the algorithms, where the Friedman test indicates the presence of significant ($p < 0.0001$) differences between the algorithms. Table \ref{tab:rules-friedman} contains the $p$-values obtained by performing the Conover post-hoc pairwise comparison procedure. First, we notice that FURIA and RIPPER generate significantly smaller rulesets than the (fuzzy)-rough rule induction algorithms. This is because both of these models use an iterative pruning process, where old rules are repeatedly replaced with more general versions. Moreover, QuickRules is by far the worst performer. One reason this might be the case, is that the iterative process of QuickRules considers the objects in the training set many times, and looks for a total ruleset that covers the entire dataset in a greedy way. Without any pruning, this is sure to deliver much more rules that algorithms with better heuristics (MODLEM), pruning (FURIA and RIPPER) and those that use the exact solutions of an optimisation problem (FRRI). Finally, let us zoom in on the contrast between FRRI and MODLEM, which is even more pronounced here. FRRI generates the smaller ruleset of the two in all but 5 of our datasets. It is interesting to observe that those datasets are highly imbalanced. Actually, this maybe considered further evidence that FRRI is better suited towards dealing with imbalanced data, as it will generate enough rules to cover both the smallest and biggest classes completely.  
For MODLEM, the worst case is vowel, with where it needs 77 more rules. On the other hand, statistical testing again shows that both MODLEM and FRRI are significantly better ($p < 0.001$) than QuickRules. Using a one-sided Wilcoxon test, we can establish that FRRI is also significantly better ($p < 0.05$) than MODLEM w.r.t.\ the size of the induced ruleset. 

\begin{table}[pht]
\centering
\begin{tabular}{llllll}
\toprule
dataset         & FRRI           & MODLEM                  & QuickRules              & FURIA                   & RIPPER                  \\ \midrule
aus             & 5.15 $\pm$ 1.5 & 2.36 $\pm$ 0.9          & 8.05 $\pm$ 1.9          & \textbf{2.17} $\pm$ 0.7 & 2.33 $\pm$ 0.7          \\
bands           & 6.29 $\pm$ 2.0 & 2.11 $\pm$ 0.6          & \textbf{1.75} $\pm$ 0.6 & 5.58 $\pm$ 1.4          & 4.75 $\pm$ 1.1          \\
bupa            & 4.20 $\pm$ 1.0 & \textbf{2.19} $\pm$ 0.6 & 4.40 $\pm$ 1.0          & 3.17 $\pm$ 0.8          & 2.40 $\pm$ 0.8          \\
cleve           & 5.45 $\pm$ 1.4 & 2.59 $\pm$ 1.0          & 8.08 $\pm$ 2.0          & 3.86 $\pm$ 0.8          & \textbf{2.50} $\pm$ 0.9 \\
derma           & 4.70 $\pm$ 2.1 & 2.56 $\pm$ 1.0          & 6.73 $\pm$ 3.5          & 2.73 $\pm$ 1.1          & \textbf{1.75} $\pm$ 0.8 \\
ecoli           & 3.85 $\pm$ 0.9 & 2.34 $\pm$ 0.8          & 3.82 $\pm$ 1.0          & 3.00 $\pm$ 0.8          & \textbf{2.00} $\pm$ 0.8 \\
glass           & 3.88 $\pm$ 1.1 & \textbf{2.10} $\pm$ 0.8 & 5.34 $\pm$ 1.5          & 3.38 $\pm$ 1.1          & 2.38 $\pm$ 0.9          \\
heart           & 4.90 $\pm$ 1.3 & 2.21 $\pm$ 0.7          & 7.58 $\pm$ 1.9          & 3.33 $\pm$ 1.0          & \textbf{1.50} $\pm$ 0.5 \\
iono            & 4.53 $\pm$ 2.2 & 1.53 $\pm$ 0.6          & 11.7 $\pm$ 5.4          & 2.71 $\pm$ 1.4          & \textbf{1.43} $\pm$ 0.5 \\
pima            & 5.08 $\pm$ 1.2 & \textbf{2.07} $\pm$ 0.7 & 5.69 $\pm$ 1.3          & 2.69 $\pm$ 1.0          & 2.25 $\pm$ 0.8          \\
sonar           & 7.80 $\pm$ 2.0 & \textbf{1.56} $\pm$ 0.5 & 21.5 $\pm$ 10           & 6.20 $\pm$ 1.4          & 4.60 $\pm$ 2.3          \\
spect           & 8.24 $\pm$ 3.2 & 1.75 $\pm$ 0.5          & \textbf{1.17} $\pm$ 0.4 & 3.06 $\pm$ 1.1          & 3.00 $\pm$ 1.4          \\
vehi            & 5.51 $\pm$ 1.4 & 2.60 $\pm$ 0.7          & \textbf{1.54} $\pm$ 0.5 & 3.81 $\pm$ 1.1          & 2.59 $\pm$ 0.8          \\
vowel           & 6.22 $\pm$ 1.5 & \textbf{2.15} $\pm$ 0.9 & 6.49 $\pm$ 1.9          & 3.74 $\pm$ 1.4          & 3.19 $\pm$ 1.3          \\
wine            & 4.51 $\pm$ 1.4 & \textbf{1.46} $\pm$ 0.5 & 6.97 $\pm$ 2.4          & 2.14 $\pm$ 0.6          & 1.60 $\pm$ 0.5          \\
red             & 5.84 $\pm$ 1.4 & \textbf{2.81} $\pm$ 0.9 & 6.76 $\pm$ 1.7          & 4.68 $\pm$ 1.3          & 3.43 $\pm$ 1.3          \\
wisc            & 3.74 $\pm$ 1.2 & 2.23 $\pm$ 0.7          & 3.84 $\pm$ 1.1          & 3.47 $\pm$ 0.7          & \textbf{1.80} $\pm$ 0.4 \\
yeast           & 4.82 $\pm$ 1.0 & 3.26 $\pm$ 1.2          & 4.93 $\pm$ 0.9          & 3.11 $\pm$ 0.9          & \textbf{2.81} $\pm$ 1.1 \\
\midrule
mean            & 5.26           & \textbf{2.21}           & 6.46                    & 3.49                    & 2.57                    \\ \bottomrule
\end{tabular}
\caption{The mean and the standard deviation of rule lengths in the rulesets of FRRI, MODLEM, QuickRules, FURIA and RIPPER on the benchmark datasets.}
\label{tab:first-atts}
\end{table}

Finally, we consider the lengths of the generated rules. Here, FRRI still performs better than QuickRules, but not significantly so, unlike MODLEM, FURIA and RIPPER, which induce significantly ($p < 0.01$) shorter rules than both QuickRules and FRRI. We examine the distribution of rule lengths across the benchmark datasets for FRRI. We find that it does not generate a few, long hyperspecific rules, which would have allowed us to easily decrease the average length by removing those rules. However, FRRI does produce rulesets with more varied rule lengths than the other algorithms, except for QuickRules, which does tend to generate long rules.

\section{Conclusions and future work}\label{sec:conclusion}

In this article, we introduced a novel rule induction algorithm, Fuzzy-Rough Rule Induction (FRRI), which combines the best parts of fuzzy and rough set theory to induce small rulesets consisting of high-accuracy rules. Our algorithm starts by discarding unnecessary information from the objects in our dataset, and then selects a minimal amount of objects to serve as rules. These rules can be viewed as summaries of the data, which are put to work to classify new samples. Experimental evaluation of the performance of FRRI in this last task showed that our algorithm is on average more accurate than state-of-the-art rule induction algorithms while using fewer rules than those that do not apply iterative pruning approaches. Moreover, the obtained rules are not much longer than those induced by state-of-the-art rough set algorithms.

We have identified some challenges for future work:
\begin{enumerate}
    \item We want to adapt our algorithm to regression and ordinal classification problems.
    \item We will explore the impact of ordering the attributes during the rule shortening phase. This can be either a single pass before the phase itself, or applying a different order for each object. Various strategies could be used, like randomizing the order of attributes, or using a discriminative ordering. Moreover, we could also use all possible minimal rules that could be created from a single object.
    \item In this version of FRRI, we were aiming at obtaining the minimal amount of rules, which is why we solved the optimisation problem exactly in Section \ref{sec:rule_selec}. Since this formulation of the problem is equivalent to the vertex cover problem from graph theory, we know it is NP-hard and cannot be solved optimally with lower complexity. However, to reduce time complexity when applying FRRI to big data, we want to explore approximate solvers, taking inspiration from the work done on approximate solvers for the vertex cover problem. Additionally, when looking at big data, we are interested in exploring a bagging ensemble approach, where we generate smaller rulesets on subsets of the data and creating one ensemble ruleset.
    \item We also want to construct hierarchical rulesets, by combining similar rules into a higher-level, more general rule, thus reducing the amount of rules and improving the explainablility.
\end{enumerate}

\section*{Acknowledgements}

Henri Bollaert would like to thank the Special Research Fund of Ghent University (BOF-UGent) for funding his research.

\bibliographystyle{plain}
\bibliography{sources}

\end{document}